\documentclass[runningheads]{llncs}
\usepackage{graphicx}
\usepackage{amssymb} 
\usepackage{float} 
\begin{document}
%
\title{Urban Region Profiling via A Multi-Graph Representation Learning Framework}
%
%
\author{
Yan Luo\inst{1}
\and
Fu-lai Chung\inst{1}
\and
Kai Chen\inst{2}
}

\authorrunning{Y. Luo et al.}

\institute{Department of Computing, The Hong Kong Polytechnic University \and
Department of Computer Science and Engineering, The Hong Kong University of Science and Technology}

\maketitle              
\begin{abstract}
Urban region profiling can benefit urban analytics. Although existing studies have made great efforts to learn urban region representation from multi-source urban data, there are still three limitations: (1) Most related methods focused merely on global-level inter-region relations while overlooking local-level geographical contextual signals and intra-region information; (2) Most previous works failed to develop an effective yet integrated fusion module which can deeply fuse multi-graph correlations; (3) State-of-the-art methods do not perform well in regions with high variance socioeconomic attributes. To address these challenges, we propose a multi-graph representative learning framework, called Region2Vec, for urban region profiling. Specifically, except that human mobility is encoded for inter-region relations, geographic neighborhood is introduced for capturing geographical contextual information while POI side information is adopted for representing intra-region information by knowledge graph. Then, graphs are used to capture accessibility, vicinity, and functionality correlations among regions. To consider the discriminative properties of multiple graphs, an encoder-decoder multi-graph fusion module is further proposed to jointly learn comprehensive representations. Experiments on real-world datasets show that Region2Vec can be employed in three applications and outperforms all state-of-the-art baselines. Particularly, Region2Vec has better performance than previous studies in regions with high variance socioeconomic attributes.

\end{abstract}
\section{Introduction}
As common constituents of the urban spaces, urban regions are spatial units consisted of building environments and socioeconomic activities conducted by people. Studying the urban region representations is significant for urban analytics due to two reasons: (1) Through profiling urban regions, the learned representations can be adopted in various downstream tasks; (2) The unified representation learning is of great importance since urban-related tasks usually depend on heterogeneous data sources. By embedding heterogeneous information into a latent space, the learned unified representations provide more knowledge for better understanding about the patterns of urban spaces, which will produce insights for urban planning, and make cities more livable and sustainable.

The facts mentioned above suggest that it is highly compelling to investigate the urban region profiling problem, which refers to embed urban regions into a latent vector space while preserving heterogeneous information. Recently, the advent of information and communication technologies leads to the proliferation of urban data. This allows researchers to explore and investigate the characteristics of urban regions via data-driven approaches. Taking human mobility (e.g., vehicle trajectories, human movement data) as an example, correlations among regions can be extracted from such activities. Remote regions may form a community since the concept "daily life circle" exists. Several existing studies \cite{RN55,RN13,RN54,RN14,RN15,RN56,RN64} indeed tried to use both POI data and human mobility to characterize region features and achieve attractive results. For instance, the method proposed in \cite{RN54} learns the region embeddings by constructing a homogeneous graph, where nodes are regions. The methods proposed in \cite{RN55,RN13} learn the region embeddings by constructing heterogeneous graphs, where nodes are a combination of urban regions, time, human mobility activities, etc. However, most of them focus on inter-region correlations. Despite promising results achieved, geographic contextual signals and region-wise inherent features have been largely overlooked in modeling urban regions. Another shortcoming of some  methods is that the multi-source heterogeneous data has not been fully fused. For example, simple concatenation used in \cite{RN14} is insufficient to extract features from multi-graphs, which typically leads to suboptimal performance in downstream tasks. A comprehensive fusion module is much desired for learning urban representation. Besides, we notice that although some state-of-the-art methods \cite{RN15,RN13} have demonstrated distinctive results in some downstream tasks, their performance in regions with high variance socioeconomic attributes is not very competitive.

To address the aforementioned challenges, this paper proposes a multi-graph, multi-source representation learning framework, called Region2Vec, to carry out highly effective urban region profiling. We notice that POI data has many inherent attributes (e.g., category, subclass). At the same time, recent research in graph embedding tends to take a graph as the input and leverage the auxiliary information to facilitate more effective embedding \cite{RN57,RN58}. Thus, we treat POI attributes as the side information and construct a knowledge graph for POIs to discover region-wise inherent properties. In addition, we also incorporate geographical neighborhood into the framework as the geographical contextual signals since adjacent regions naturally show direct correlations according to the First Law of Geography \cite{RN24}. By incorporating multi-source data including global-level human mobility, local-level geographical neighborhood, and region-wise POI side information, the urban spaces are comprehensively depicted by "man-land-dynamic-static", which are the classical four dimensions in Geography \cite{RN62}. Then, through mobility pattern similarity analysis, topology analysis, and constructing knowledge graph, we can then encode accessibility, vicinity, functionality correlations among regions by constructing graphs. To better propagate information for every single data source, the graph attention network is employed. Furthermore, to promote cooperation of different graph representations, a multi-graph fusion module with some designed learning objectives is proposed to model the underlying correlations among graphs in a joint manner. The contributions of this paper can be summarized as follows.

\begin{itemize}
\item A novel region embedding framework for urban region profiling is proposed. The learned region representations preserve global-level inter-region correlations, local-level geographical contextual signals, and inherent region-wise attributes.
\item A multi-graph fusion module is devised to integrate multiple graphs. It is capable of fusing multi-source urban data into comprehensive latent representations, with the collaboration of global encoder and accessibility/ vicinity/ functionality correlation decoder.
\item Comprehensive experiments on real-world datasets were conducted to demonstrate the distinctive performance of Region2Vec. In particular, Region2Vec has better performance than previous studies in regions with large variance socioeconomic attributes.
\end{itemize}

\section{Related Work and Preliminaries}

The goal of multi-source heterogeneous data embedding is to map multiple source information into a continuous low-dimensional latent feature space. 
Complementary information from multiple sources will generate features that cannot be captured by the individual source \cite{RN4}. Many studies demonstrate that multi-source heterogeneous data embedding deepens the fusion of information from different sources and achieves better performance than the state-of-art uni-source methods \cite{RN45,RN44,RN46}. For our problem setting, we focus on multi-source data fusion based on multi-graphs, which has been mainly based on cross-source regularization \cite{RN52,RN51}. However, these studies were not tailored for urban scenarios. Before introducing our method for urban regional profiling, let us give some definitions and present the problem statement.


\begin{definition}
\textbf{Human Mobility} can be defined as a set of trips conducted by citizens in urban spaces. A trip in human mobility datasets starts from an origin point (O) and ends at a destination point (D). Thus, a trip can also be named an OD. If we link the O/D of a trip with the urban regions to which they belong, then we can denote human mobility dataset $M$ as:
\begin{equation}
\label{eq:1}
M=\left\{\overrightarrow{m_{0}}, \overrightarrow{m_{1}}, \ldots, \overrightarrow{m_{|M|}}\right\}, \overrightarrow{m_{\cdot}}=\left(r_{o}, r_{d}\right)
\end{equation}
where $\overrightarrow{m_{\cdot}}$ is a trip that can be represented as a two dimensional vector. $r_o$ is the origin region and $r_d$ is the destination region. $|M|$ represents the length of $M$.
\end{definition}

\begin{figure}[htb]
\centering
\includegraphics[width=0.68\textwidth]{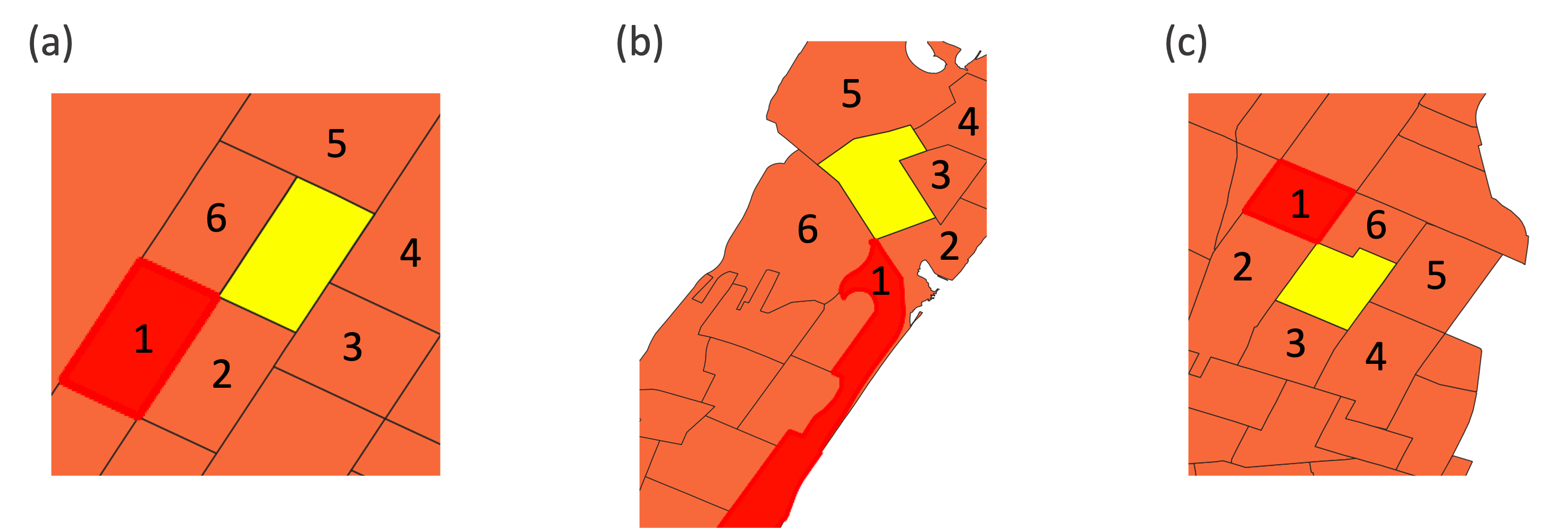} 
\caption{Three examples of geographic neighborhood. Yellow areas are the chosen areas. Their geographic neighborhoods are numbered, while diagonal neighborhoods are all numbered as 1.}
\label{fig2}
\end{figure}

\begin{definition}
\textbf{Geographic Neighborhood} of a region is described based on spatial adjacency. 
If two regions have pixels connected, then they are geographic neighbors to each other. It is worth noting that the number of geographic neighborhoods of different regions may be different due to the irregular shapes of urban regions. Examples are given in Figure \ref{fig2}.

For each region $r_i$, we can get a vector of variable length via traversing all the urban regions. Supposing that there are $N$ urban regions in total, the geographic neighborhood set can be denoted as:
\begin{equation}
\label{eq:2}
N=\overrightarrow{n_{0}}, \overrightarrow{n_{1}}, \ldots, \overrightarrow{n_{N}}, \overrightarrow{n_i}=\left(r_{1}, r_{2}, \ldots, r_{|\overrightarrow{n_i}|}\right)
\end{equation}
where $\overrightarrow{n_i}$ is a geographic neighborhood vector for an urban region $r_i$; $r_{\cdot}$ is a neighborhood for the urban region. $|\overrightarrow{n_i}|$ represents the length of $\overrightarrow{n_i}$.
\end{definition}

\begin{definition}
\textbf{POI Side Information} refers to different attributes of POI. Since POIs are the direct representations of urban functions, features from POI side information can be regarded as meta-knowledge, which reflects region's functional attributes. Side information can help to establish correlations among those POIs, and also to model relations among urban regions. Firstly, we map POIs to the located region. Then, POI side information dataset can be denoted as follows:
\begin{equation}
\label{eq:3}
S=\overrightarrow{s_{0}}, \overrightarrow{s_{1}}, \ldots, \overrightarrow{s_{N}}, \overrightarrow{s_i}=\left(s_{1}, s_{2}, \ldots, s_{|\overrightarrow{s_i}|}\right)
\end{equation}
where $\overrightarrow{s_i}$ is a POI side information vector for a urban region $r_i$. $s_{\cdot}$ is a kind of POI attribute.
\end{definition}


\noindent\textbf{Problem Statement (Urban Region Embedding).} \noindent Given three sets of vectors $M$, $N$, and $S$ described by Equation \ref{eq:1}, \ref{eq:2}, and \ref{eq:3} respectively, this research aims to learn a distributed and low dimensional embedding $v_i$ for each urban region $r_i$. The embedding set can be denoted as:
\begin{equation}
\label{eq:4}
V=\overrightarrow{v_{0}}, \overrightarrow{v_{1}}, \ldots, \overrightarrow{v_{N}}, v_{i} \in \mathbb{R}^{d}
\end{equation}
where $d$ is the uniform dimension.
The embedding set $V$ should preserve as much information of human mobility, geographic neighborhood, and POI side information as possible.

\section{Region2Vec}
In this section, we introduce the proposed multi-graph, multi-source representation learning framework, namely Region2Vec, for urban region embedding. We first present an overview of the framework. Then, we elaborate the three main modules in our framework, namely, correlation modeling, graph attention network (GAT) \cite{RN19} and multi-graph fusion.

\subsection{Framework Overview}
\begin{figure}[htb]
\centering
\includegraphics[width=0.95\textwidth]{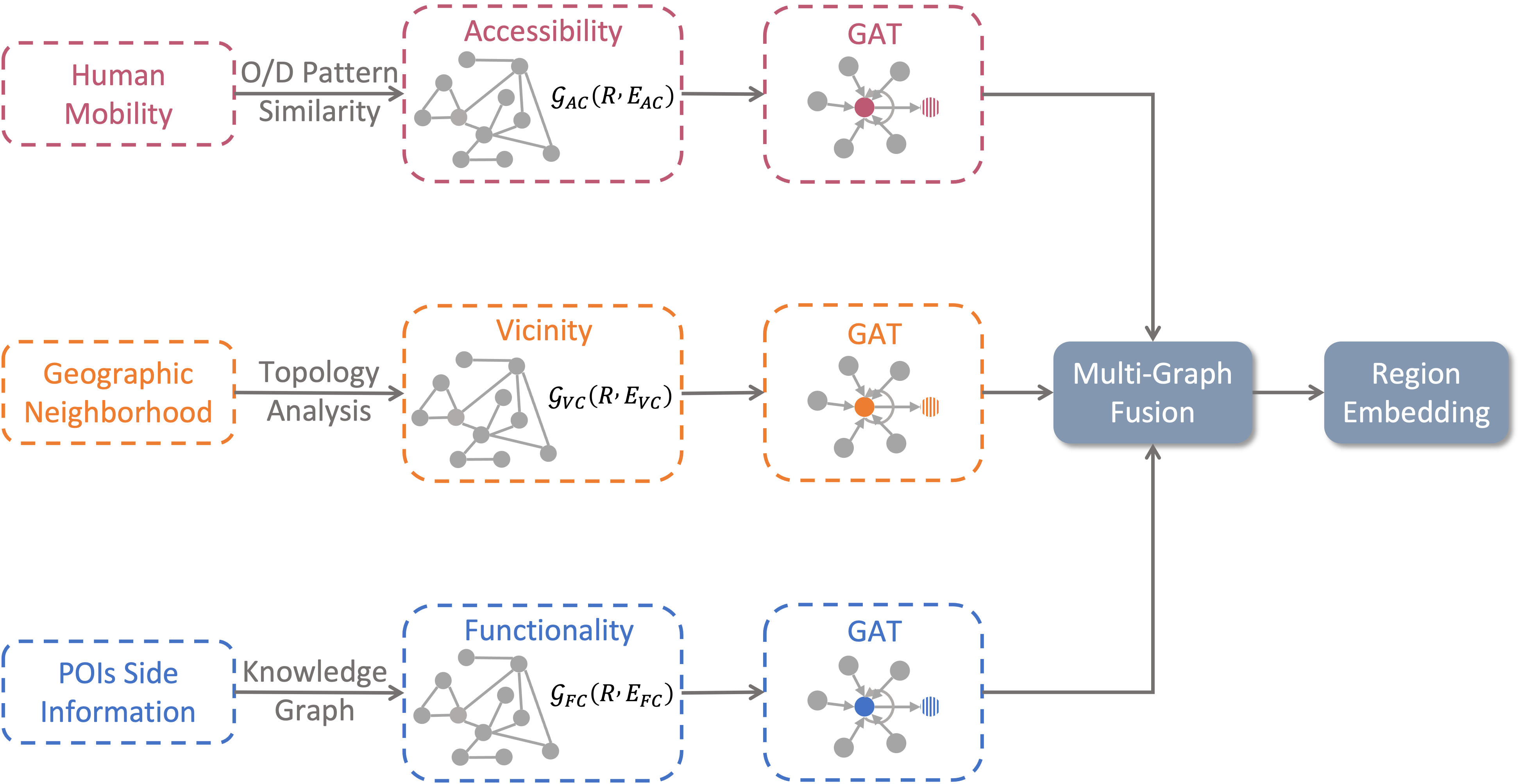} 
\caption{Processing pipeline of the proposed multi-graph, multi-source representation learning framework Region2Vec.}
\label{fig3}
\end{figure}

Figure \ref{fig3} shows the pipeline of our proposed multi-graph, multi-source representation learning framework. Different sources of urban data, including human mobility, geographic neighborhood, and POI side information can be encoded via multiple graphs. First, a correlation modeling module is introduced to construct multiple graphs based on multi-source data. Then, a graph attention network is used to aggregate and update information in each graph. After that, a multi-graph fusion module is proposed to deeply integrate multi-graph information. In this way, the final embedding incorporates non-Euclidean correlations among regions based on human mobility, geographic neighborhood, and POI side information.

\subsection{Correlation Modeling Module}
Correlations among urban regions can be characterized by different aspects. From the aspect of human mobility, a trip has an origin region and a destination region, which can form a correlation. For many trips, origin regions/destination regions can also be related to other origin regions/destination regions based on mobility patterns in terms of accessibility. As for geographic neighborhood, vicinity in space can be revealed. As the representation of urban functions, POI side information reflects functionality correlations. Similar regions in terms of accessibility, vicinity, and functionality will show high correlations due to proximity in non-Euclidean space. In our study, we construct three types of region correlations based on human mobility, geographic neighborhood, and POI side information as follows.

\subsubsection{Accessibility Correlation (AC) Based on Human Mobility.} Human mobility directly reveals the inter-region interaction movement between people and urban spaces. It is found that if trips have the same O/D regions, then different D/O regions of trips are similar \cite{RN13}. In other words, through the O/D pattern similarity, important underlying accessibility correlation can be modeled and captured based on human mobility. Suppose we have a human mobility dataset $M$, the similarity value between region $r_i$ and region $r_j$ is computed as:
\begin{equation}
\label{eq:5}
s_{r_{j}}^{r_{i}}=\left|\left(r_{i}, r_{j}\right) \in M\right|
\end{equation}
where ($r_i$,$r_j$) formed a trip in $M$, and $|.|$ calculates its length. In other words, the similarity value between two regions is actual the count of co-occurrence of these two regions. Then, the O/D pattern similarity and accessibility correlations among regions can be defined as:
\begin{small}
\begin{equation}
\label{eq:6}
p_{o}\left(r \mid r_{i}\right)=\frac{s_{r_{i}}^{r}}{\sum_{r} s_{r_{i}}^{r}}, p_{d}\left(r \mid r_{i}\right)=\frac{s_{r}^{r_{i}}}{\sum_{r} s_{r}^{r_{i}}}
\end{equation}
\end{small}
\begin{small}
\begin{equation}
\label{eq:7}
AC_{o}^{i j}={simi}\left(p_{o}\left(r \mid r_{i}\right), p_{o}\left(r \mid r_{j}\right)\right), AC_{d}^{i j}={simi}\left(p_{d}\left(r \mid r_{i}\right), p_{d}\left(r \mid r_{j}\right)\right)
\end{equation}
\end{small}
where $simi(.)$ is the function for calculating the cosine similarity; $r$ denotes a certain region; $A C_{o}^{i j}$ is the accessibility correlation between two O regions; $A C_{d}^{i j}$ denotes the accessibility correlation between two D regions.

\subsubsection{Vicinity Correlation (VC) Based on Geographic Neighborhood.}

Spatial vicinity is a kind of important correlation due to the First Law of Geography \cite{RN24}. Adjacent regions in space are naturally more similar. According to Equation \ref{eq:2}, through topology analysis, the geographic neighborhood dataset $N$ contains geographic neighborhood vector $\overrightarrow{n_i}$ for each urban region $r_i$. These vectors actually represent vicinity correlations among regions. Hence, the vicinity correlations can be described as:
\begin{equation}
\label{eq:9}
V C^{i j}={simi}\left(\overrightarrow{n_{i}}, \overrightarrow{n_{j}}\right)
\end{equation}
where $VC^{i j}$ is the vicinity correlation between region $r_i$ and $r_j$.

\subsubsection{Functionality Correlation (FC) Based on POI Side Information.}
The POI side information of a region reveals the functionality and also reflects intra-region features. To include more accurate, diverse, and explainable information, it is necessary to go beyond POI itself and take the POI attributes (i.e., POI side information) into account. We here choose to use the Knowledge Graph (KG) of POI side information to construct the functionality correlation model.

\begin{figure}[htb]
\centering
\includegraphics[width=\textwidth]{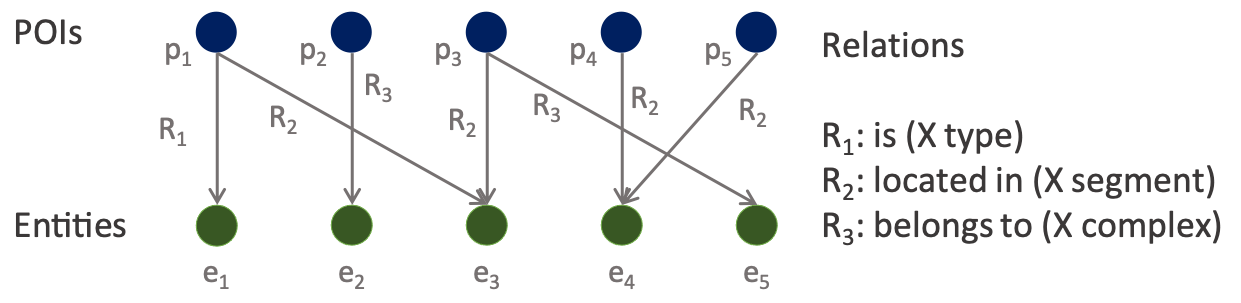} 
\caption{A toy example for constructing knowledge graph. We list only three relations here.
}
\label{fig4}
\end{figure}

When constructing KG for POI and its side information, POIs and side attributes can be regarded as nodes, while different types of relations among nodes are regarded as different types of edges. We give a toy instance in Figure \ref{fig4}. Here, we list only three types of relations. Some simple connectivities can be easily found such as $p_{1}-R_{2}-e_{3}$. However, if we further treat the edges as reversible, we can capture the long-range connectivity in the knowledge graph, such as $p_{1}-R_{2}-e_{3}-\left(-R_{2}\right)-p_{3}-R_{3}-e_{5}$, to learn richer information from KG. Thus, reversible edges are adopted when constructing KG.

In this paper, TransD \cite{RN39} is employed for getting functionality region embeddings  $\overrightarrow{s_i}$ through KG. The functionality correlations are described as:
\begin{equation}
\label{eq:10}
F C^{i j}={simi}\left(\overrightarrow{s_{i}}, \overrightarrow{s_{j}}\right)
\end{equation}
where $FC^{i j}$ is the functionality correlation between region $r_i$ and $r_j$.

\subsection{Graph Attention Network Module}

We construct graphs for accessibility correlation $AC$, vicinity correlation $VC$, and functionality correlation $FC$, respectively. Nodes in graphs are connected with $k$ nearest neighbors of them. Each graph can be denoted as $\mathcal{G}(R, C)$, where $R=\left\{r_{i}\right\}_{i=1}^{n}$  represents $n$ regions and $C=\left\{C_{i}\right\}_{i=1}^{k}$ denotes the edges. Therefore, we have $\mathcal{G_{AC}}$, $\mathcal{G_{VC}}$, and $\mathcal{G_{FC}}$ based on different kinds of correlations. Then, the GAT is applied to integrate and update node representations of each graph. The attention mechanism on the graph-structured data can automatically learn weights of information from neighbors of a node during the propagation. The output node representation of the GAT module of each graph can be denoted as $E_{AC}$, $E_{VC}$, and $E_{FC}$, respectively. The output node representation contains information of its neighbors.

\subsection{Multi-Graph Fusion Module}

\begin{figure}[htb]
\centering
\includegraphics[width=\textwidth]{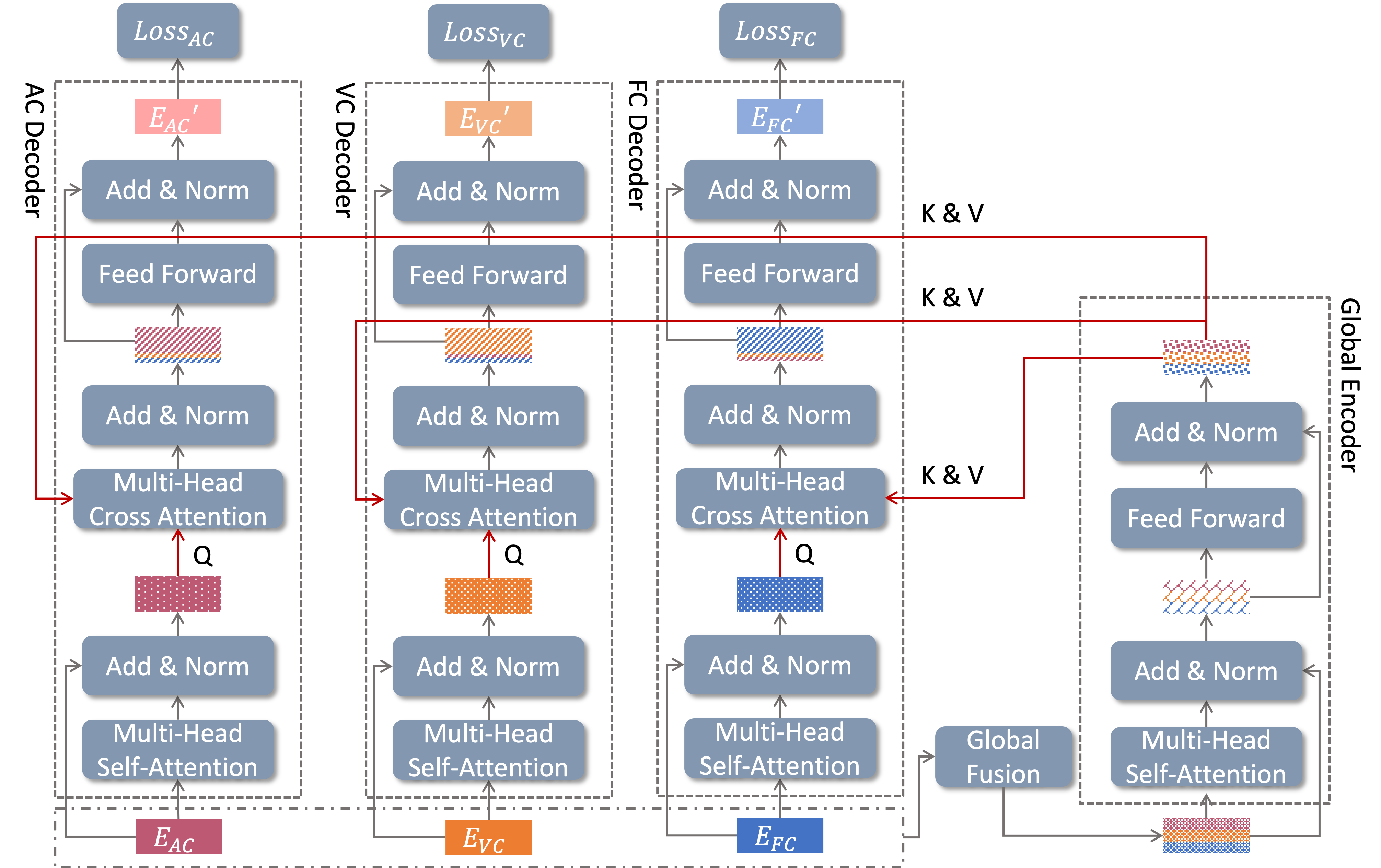} 
\caption{Architecture of multi-graph fusion module}
\label{fig5}
\end{figure}

With the GAT modules integrating and updating different types of region information for each graph, a multi-graph fusion module is exploited to fuse all types of information into the final region embeddings. Region correlations from different graphs are highly related. Taking human mobility and POI side information as an example, similar OD pairs in the morning peak and evening peak generally represent commuting between residence districts and business districts. As for geographical neighborhood and human mobility/POI side information, we can use the First Law of Geography to explain it — "Everything is related to everything else, but near things are more related to each other". Such relationships among these three graphs give us an intuition that incorporating information from multiple graphs will not only improve the performance but also enhance the learning process for each graph. A multi-graph fusion module can endow the Region2Vec with the capability to incorporate spatial semantics from region-wise side information, local-level geographical adjacent relations, and global-level mobility pattern. As shown in Figure \ref{fig5}, we employ an encode-decode architecture to facilitate effective integration among multiple graphs. Then, we design a loss function as our overall learning objective.

\subsubsection{Global Encoder.}

Multi-graph representations $E_{AC}$, $E_{VC}$, and $E_{FC}$ are first concatenated and fed into the global fusion layer, which generates comprehensive region embeddings by a single layer Multi-Layer Perceptron (MLP). The fused representation $E_f$ can be denoted as:
\begin{small}
\begin{equation}
\label{eq:11}
E_f = \sum{\sigma\left(E_mW+b\right)E_m}, m \in \{AC, VC, FC\}
\end{equation}
\end{small}
where $W$ and $b$ are learnable parameters in the MLP.

Following the transformer architecture \cite{RN61}, the global encoder layer makes use of two sub-layers, namely, a multi-head self-attention mechanism for enabling further integration of information and a fully connected feed-forward network for deep feature extraction. Residual connection and layer normalization are employed around each of these two sub-layers.

\subsubsection{AC/VC/FC Decoder.}

In addition to the two sub-layers mentioned in the encoder layer, the AC/FC/VC decoder applies a multi-head cross attention, which treats the output of the encoder layer as the key and value, and uses their attention result as the query. The reason we use multi-head cross attention mechanism rather than multi-head self-attention in \cite{RN15} is that, such an attention mechanism can provide a latent adaptation across AC/FC/VC, and the global fusion result. When three sources of data are integrated in global fusion part, there will be an information bottleneck due to information compression. Cross attention can reduce the noise brought by global fusion result to AC/FC/VC. For the result $E_m\prime$ via the self-attention mechanism for each graph representation $E_m$, the query matrix $Q\in\mathbb{R}^{n\times d}$, key matrix $K\in\mathbb{R}^{n\times d}$ and value matrix $V\in\mathbb{R}^{n\times d}$ can be defined as:
\begin{small}
\begin{equation}
\label{eq:12}
Q=E_fW_Q,K=E_m\prime W_K,V=E_m\prime W_V
\end{equation}
\end{small}
where the projections are carried out by parameter matrices $W_Q$, $W_K$, and $W_V$. Residual connection and layer normalization are also employed around each of the three sub-layers.


\subsubsection{Loss Function Designation.}

Through the multi-graph fusion module, features of each graph are updated. Various learning tasks are then devised based on these updated graph representations. The overall learning objective can be represented as:
\begin{small}
\begin{equation}
\label{eg:13}
\mathcal{L}={Loss}_{A C}+{Loss}_{V C}+{Loss}_{F C}
\end{equation}
\end{small}
where ${Loss}_{AC}$, ${Loss}_{VC}$, and ${Loss}_{FC}$ are losses for AC, VC, and FC reconstruction respectively.

Here, AC is reconstructed by maximizing the probability of O/D occurrence. We expect that the possibility of predicting the O/D region given the D/O region based on the region representations will be the highest. Hence, the AC reconstruction loss between region $r_i$ and region $r_j$ can be computed as:


\begin{small}
\begin{equation}
\hat{p}_{o} \left(r_{j} \mid r_{i} \right) =\frac{\exp \left(E_{o}^{i^{T}} E_{d}^{j}\right)}{\sum_{j} \exp \left(E_{o}^{i^{T}} E_{d}^{j}\right)},\hat{p}_{d} \left(r_{j} \mid r_{i} \right) =\frac{\exp \left(E_{d}^{i^{T}} E_{o}^{j}\right)}{\sum_{j} \exp \left(E_{d}^{i^{T}} E_{o}^{j}\right)}
\end{equation}
\end{small}

\begin{small}
\begin{equation}
\label{eq:15}
{Loss}_{AC} = \sum_{\left(r_i,r_j\right)\in\mathcal{M}}{-log{\hat{p}}_o\left(r_j\middle| r_i\right)-log{\hat{p}}_d\left(r_j\middle| r_i\right)}.
\end{equation}
\end{small}


On the other hand, the VC/FC reconstruction loss is designed to make the final region representation preserve the information from geographical neighborhood and POI side information. It can be computed as:
\begin{small}
\begin{equation}
\label{eq:16}
{Loss}_{VC}=\sum_{i, j}\left(VC^{ij}-E_{VC}^{i} E_{VC}^{j}\right)^{2}, {Loss}_{FC}=\sum_{i, j}\left(FC^{ij}-E_{FC}^{i} E_{FC}^{j}\right)^{2}.
\end{equation}
\end{small}

\section{Experiments}

In this section, we report the results of several experiments conducted on real-world datasets to evaluate the performance of Region2Vec. The design of our experiments is to answer the following questions:

\noindent\textbf{RQ1:} How well does Region2Vec perform in various downstream urban analytics tasks?

\noindent\textbf{RQ2:} How do different modules of Region2Vec contribute to the model performance?
%

\subsection{Study Area and Datasets}

We chose Manhattan borough in New York City as our study area, which is also the benchmark dataset in this research field. The study area is divided into 180 regions. From NYC Open Data\footnote{
http://opendata.cityofnewyork.us/}, we use real-world datasets such as census block shapefile, taxi trips, POI data, 
and check-in data. We also find the district division provided by the community boards from \cite{RN9}. 
The details of these datasets can be found in Table \ref{table2}. The description of POI data is shown in Table \ref{table3}.

\begin{table}[htb]
\caption{Details of dataset.}
\centering
\resizebox{\columnwidth}{!}{
\begin{tabular}{l|l}
\hline
Dataset & Details\\
\hline
Census blocks & 180 block boundaries split by streets in Manhattan. \\
Taxi trips & $\sim$ 10 million taxi trip records accumulated in one month. \\
POI data & $\sim$ 6 thousand POIs (including 9 types of attributes).\\
Check-in data & $\sim$ 80 thousand check-in records during one year. \\
District division & Manhattan is divided into 12 districts based on land usage. \\ 
\hline
\end{tabular}}
\label{table2}
\end{table}


\begin{table}[htb]
\caption{Field description of POI dataset}
\centering
\resizebox{\columnwidth}{!}{
\begin{tabular}{l|l}
\hline
Attribute field & Description                                                                                                        \\ \hline
PLACEID         & The unique identifier for each POI.                                                                                \\
SOURCE          & Agency that defined the POI.                                                                                       \\
FACILITY\_T     & Categories of POIs.                                                                                                \\
FACI\_DOM       & Subclasses of POIs.                                                                                                \\
SEGMENTID       & POI is assigned the closest roadbed SEGMENTID.                                                                     \\
PRI\_ADD        & POI has PRI\_ADD field if the POI is related to any address point.                                                 \\
BIN             & Point is assigned a Building Identification Number (BIN) if it falls within a building. \\
SOS             & Indicates which side of the street the POI is on.                                                                  \\
SAFTYPE         & Point is assigned a SAFTYPE if it is a part of a Complex.                                                          \\
COMPLEXID       & Point is assigned a COMPLEXID if it is a part of a Complex.                                                        \\ \hline
\end{tabular}}
\label{table3}
\end{table}

\subsection{Downstream Tasks for Evaluation}
\subsubsection{Region Clustering Visualization.} Regions may fall into the same category if their land-use type is similar. To verify whether our obtained region embeddings effectively fuse multiple graphs to contain the information of land use, we cluster region embeddings using K-means and visualize the results to intuitively interpret them. As shown in Figure \ref{fig6} (a), the district division data from community boards \cite{RN9} which divide Manhattan into 12 components are used as ground truth. Thus, we partition the study area into 12 clusters. For the clustering result, regions with the same land use type should be in the same group.

\subsubsection{Region Clustering Evaluation.} We adopt Normalized Mutual Information (NMI) and Adjusted Rand Index (ARI) to further quantitively evaluate region clustering results of the proposed embedding method and baselines.

\subsubsection{Popularity Prediction.} Check-in volume is regarded as the popularity. 
Lasso regression model \cite{RN20} is employed in this task to predict the popularity. The independent variables are region embeddings, while the dependent variable is the popularity. In this part, we use three metrics: Mean Absolute Error (MAE), Root Mean Square Error (RMSE), and the coefficient of determination ($R^2$), to measure the performance of different approaches. The first two metrics are to quantify the errors of prediction, while the last one is to estimate the goodness of fit of the models. We measure the performance using these three metrics by K-Fold cross-validation, where K is set as 5. The value of the L1 normalization weight of Lasso regression is chosen by grid searching.

\subsection{Performance Comparison (RQ1)}

We adopted the following seven approaches to benchmark the performance of our proposed Region2Vec:

\noindent\textbf{(I) Graph Embedding Baselines}\\
\begin{itemize}
    \item \textbf{GAE:} We apply Graph Auto-Encoder (GAE) proposed in \cite{RN10} on multi-graphs and make the GAE model of each graph share the middle layer to learn region embeddings.
    \item \textbf{DeepWalk:} We apply the DeepWalk model \cite{RN60} on multi-graph data and concatenate the embeddings of each graph to get region embeddings.
    \item \textbf{LINE:} We apply the LINE model \cite{RN11} on multi-graph data and concatenate the embeddings of each graph to get region embeddings.
    \item \textbf{Node2Vec:} We apply node2vec \cite{RN12} on multi-graph data and concatenate the embeddings of each graph to get region embeddings.
\end{itemize}

\noindent\textbf{(II) State-of-the-Art Methods}\\
\begin{itemize}
    \item \textbf{ZE-Mob:} ZE-Mob proposed in \cite{RN13} learns region embeddings by considering the co-occurrence relation of regions in human mobility.
    \item \textbf{MV-PN:} MV-PN proposed in \cite{RN14} learns region embeddings with a region-wise multi-view POI network.
    \item \textbf{MV-Embedding:} MV-Embedding proposed in \cite{RN15} learns region embeddings based on both human mobility and inherent region properties.
\end{itemize}

In the experiments, the embedding size of ZE-Mob is set to 96 as recommended by the authors. Thus, the embedding sizes of baselines, our model, and variants are all set as 96. To enhance the GAT performance, an 8-head attention mechanism is employed in each GAT layer. For the multi-graph fusion module, we set the head number $h$ of multi-head self-attention as 4.

\subsubsection{Region Clustering Visualization.}
\begin{figure}[htb]
\centering
\includegraphics[width=\textwidth]{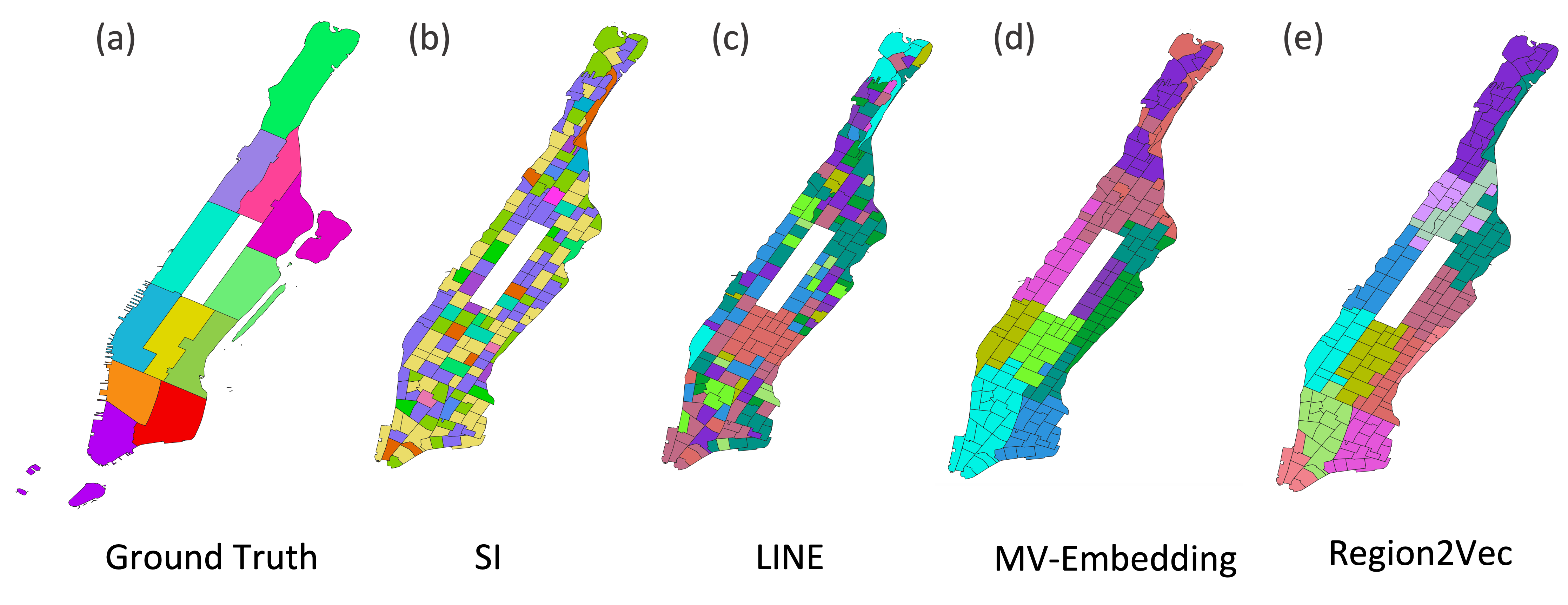} 
\caption{Region clustering results for some methods in Manhattan borough. (a) Ground Truth; (b) SI; (c) LINE; (d) MV-Embedding; (e) Region2Vec. The same color marks regions in the same cluster.}
\label{fig6}
\end{figure}

We can see from the comparison between (c)-(e) and (b) in Figure \ref{fig6} that multi-graph methods are much better than the uni-graph method. In addition, our proposed Region2Vec has produced more ideal clustering than another state-of-the-art method MV-Embedding, in terms of consistency with real boundaries of ground truth. Another interesting finding is that in Manhattan uptown, which is a typical area with high variance socioeconomic attributes in Manhattan, Region2Vec has shown distinguished performance compared with others. This may be due to two reasons: (1) Exploiting POI side information by KG can greatly mine the socioeconomic-related knowledge; (2) Our multi-graph fusion module can deeply fuse the information from different sources as compared with other methods.

\subsubsection{Region Clustering Evaluation}

\begin{figure}[htb]
\centering
\includegraphics[width=0.8\textwidth]{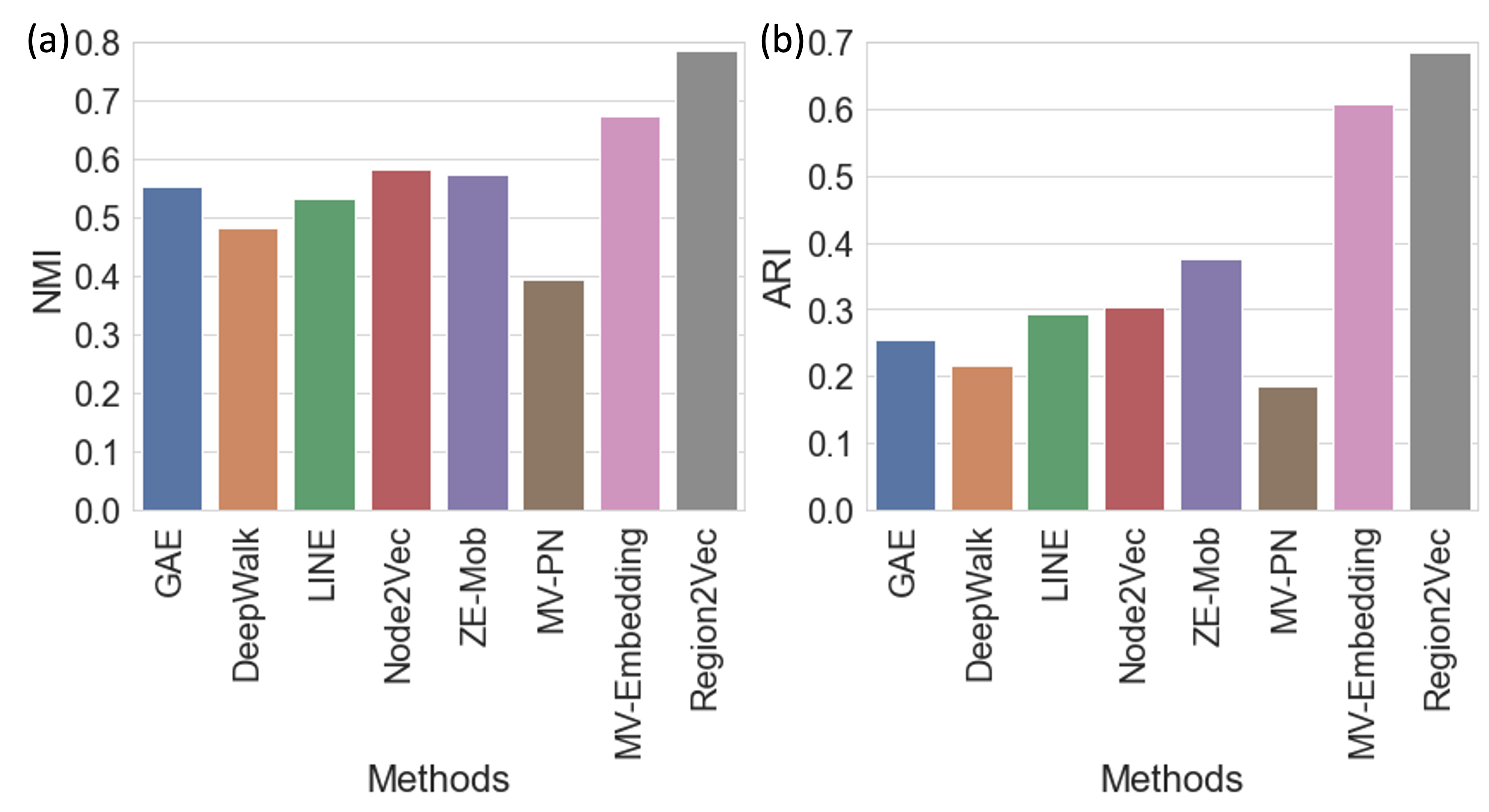} 
\caption{Results of region clustering obtained by different baselines and state-of-the-art methods.}
\label{fig7}
\end{figure}

Figure \ref{fig7} shows the NMI and ARI of region clustering results obtained by all approaches. We can see that: (1) Region2Vec outperforms all baseline approaches. Compared with the state-of-the-art methods, it has a 16.64\% increase in performance in NMI and a 12.66\% increase in performance in ARI. (2) Methods for a simple combination of multi-graphs, such as GAE, LINE, and Node2Vec, obviously cannot make full use of multi-graph information.

\begin{figure}[htb]
\centering
\includegraphics[width=0.95\textwidth]{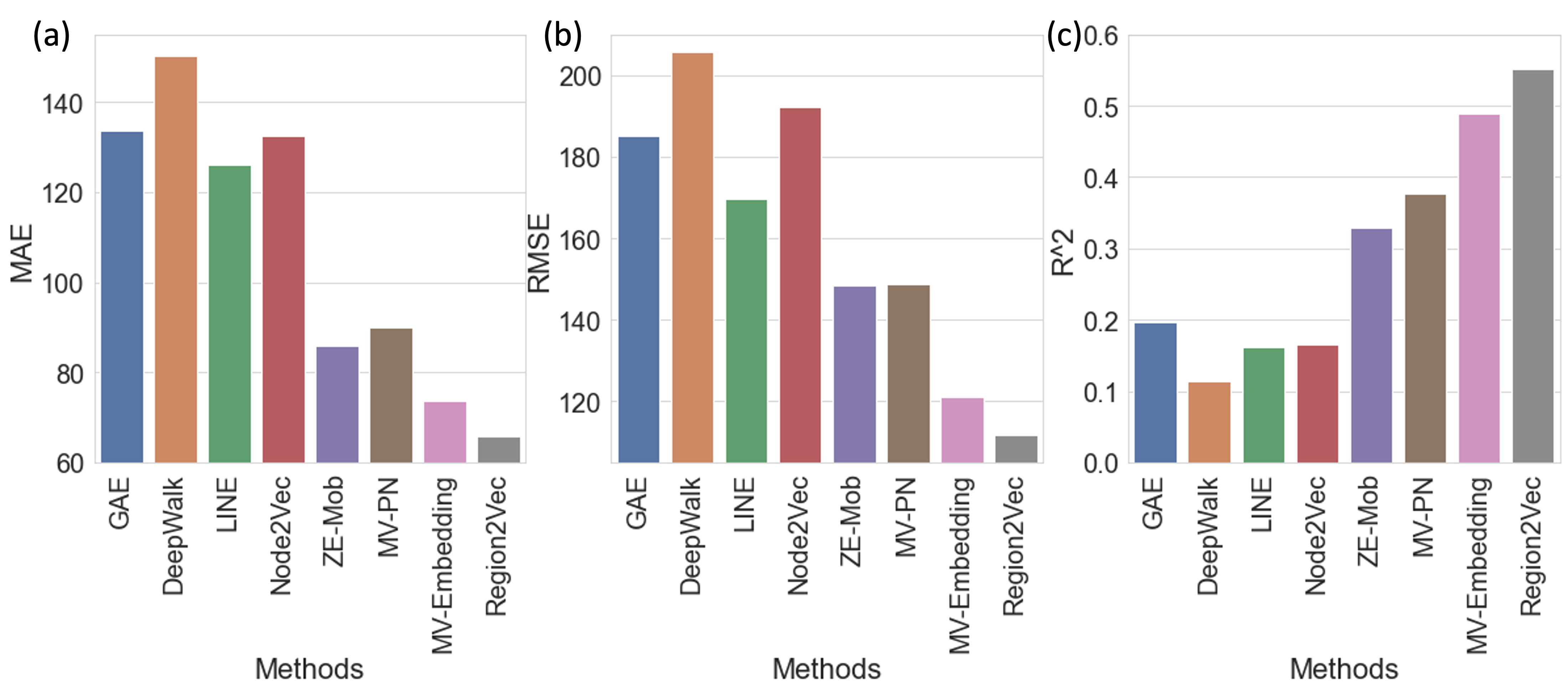} 
\caption{Results of popularity prediction obtained by different baselines and state-of-the-art methods.}
\label{fig9}
\end{figure}

\subsubsection{Popularity Prediction}

The experimental results of popularity prediction are presented in Figure \ref{fig9}. Compared with the state-of-the-art methods, Region2Vec has 10.34\%, 7.80\%, and 12.88\% improvement in MAE, RMSE, and $R^2$, respectively.

\subsection{Ablation Study (RQ2)}

To better understand the effect of each module in the multi-graph multi-task training of Region2Vec, we perform the following ablation experiments:

\noindent\textbf{(I) Ablation Study for Correlation Modeling}



\begin{itemize}
    \item \textbf{HM/GN/SI:} Region2Vec applied merely on HM/GN/SI.
    \item \textbf{HM+GN:} Region2Vec applied merely on HM and GN.
    \item \textbf{HM+SI:} Region2Vec applied merely on HM and SI.
\end{itemize}



\noindent\textbf{(II) Ablation Study for GAT Module}

\begin{itemize}
    \item \textbf{R2V-g:} Region2Vec without GAT module to propagate and update information for each graph.
\end{itemize}

\noindent\textbf{(III) Ablation Study for Multi-Graph Fusion Module}
\begin{itemize}
    \item \textbf{R2V-f:} Region2Vec without multi-graph fusion module to deeply integrate all graphs. Every graph is assigned equal weights when obtaining the final region embedding result.
    \item \textbf{R2V-m:} We use the fusion module in MV-embedding while disabling the multi-graph fusion module for Region2Vec.
\end{itemize}


\subsubsection{Region Clustering Visualization.} We can see from the comparison between (c)-(e) and (b) in Figure \ref{fig6} that multi-graph methods are much better than the uni-graph methods.

\begin{figure}[H]
\centering
\includegraphics[width=0.8\textwidth]{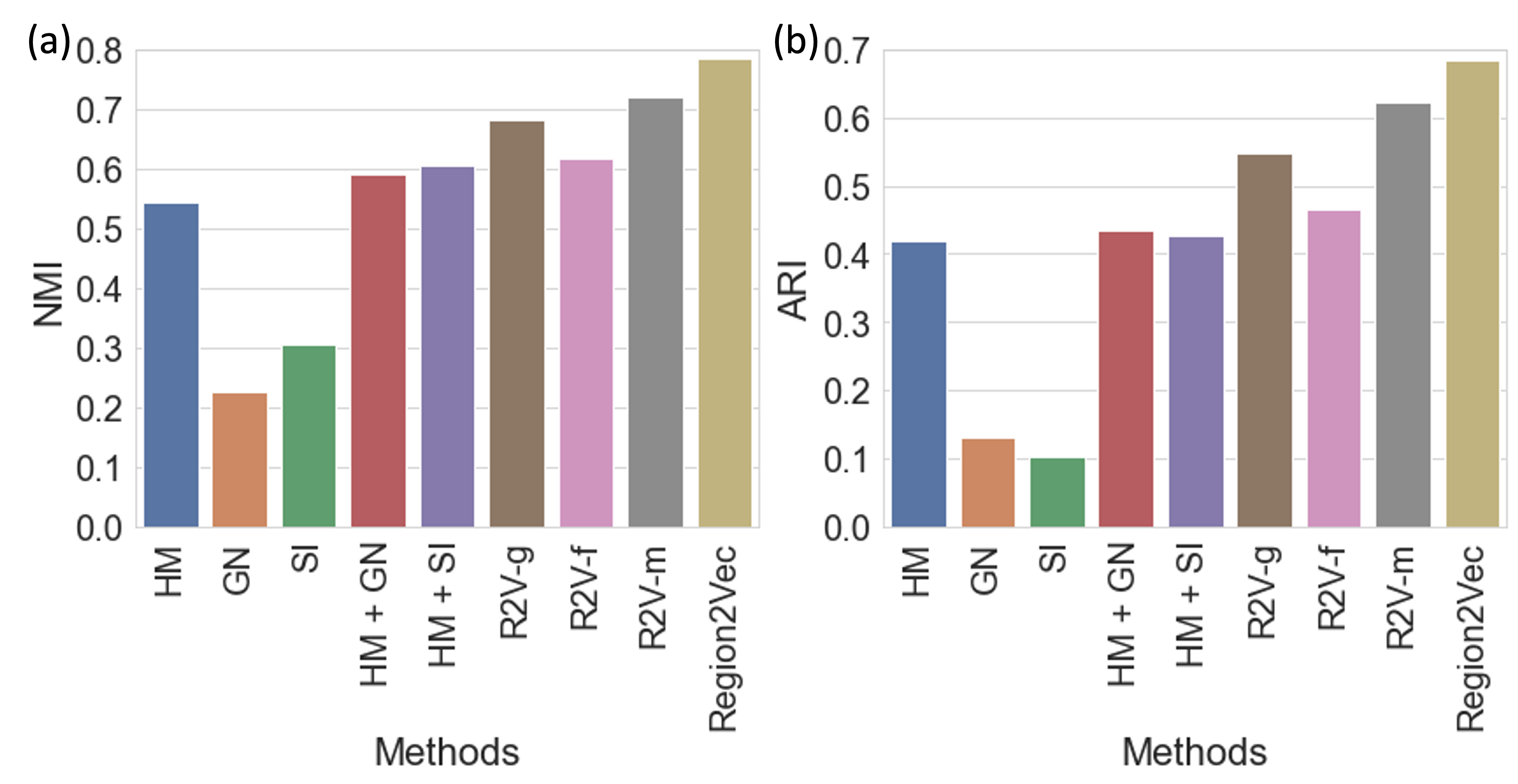} 
\caption{Results of region clustering evaluation using variants of Region2Vec.}
\label{fig10}
\end{figure}

\subsubsection{Region Clustering Evaluation.} We can see from Figure \ref{fig10} that: (1) Every module in Region2Vec is necessary. (2) The methods for bi-graph (i.e., HM+GN/HM+SI) generally have better performance than the methods for uni-graph (i.e., HM/GN/SI). Among uni-graph methods, HM marginally outperforms the other two methods, which indicates the importance of accessibility correlation in region clustering tasks. (3) If we just simply combine the information from multi-graph (i.e., R2V-g, R2V-f, and R2V-m), it will cause at least 8.73\% and 9.78\% reduction of performance in NMI, and ARI, respectively.



\begin{figure}[H]
\centering
\includegraphics[width=\textwidth]{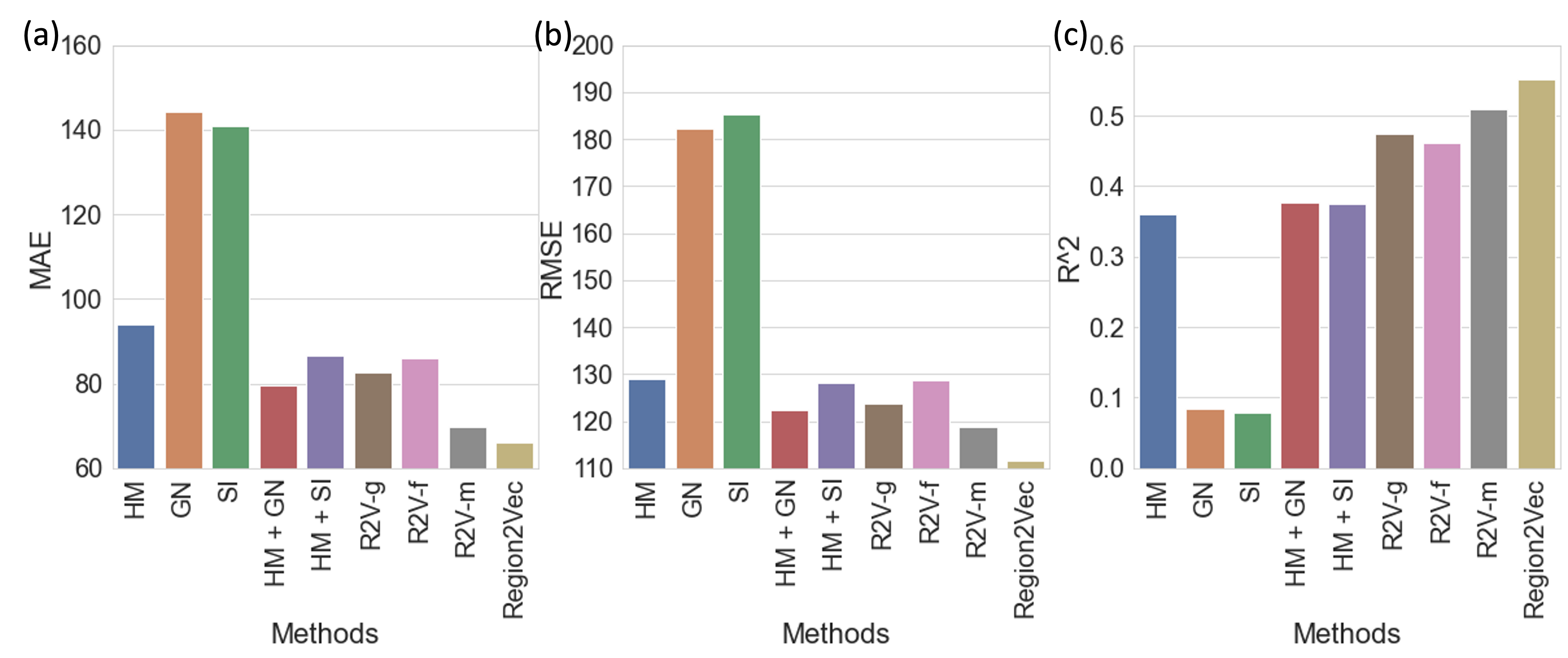} 
\caption{Results of popularity prediction using variants of Region2Vec.}
\label{fig12}
\end{figure}

\subsubsection{Popularity Prediction.} Results are presented in Figure \ref{fig12}. Region2Vec has at least 5.64\%, 6.11\%, and 8.45\% improvement in MAE, RMSE, and $R^2$, respectively. The necessity of each module is soundly verified.




\section{Conclusion}
In this work, we study the urban region profiling problem. Region2Vec, a multi-graph, multi-source representation learning framework, is proposed to learn comprehensive embeddings for urban regions. Through urban region profiling, the task-agnostic framework is capable of handling various applications. In particular, multi-source urban data
are encoded using multiple graphs for representing inter-region relations, geographical contextual information, and intra-region information.
With the GAT module employed to aggregate and update information in each graph, a multi-graph fusion module is devised to jointly learn comprehensive representations. Experiments on real-world datasets demonstrate that Region2Vec outperforms all state-of-the-art baselines in three downstream tasks. Our future work includes making our framework more task-oriented and focusing on the interpretability of our model.


%
%
\bibliographystyle{splncs04}
\bibliography{aaai22}

\end{document}